# Adaptive Homophily Clustering: Structure Homophily Graph Learning with Adaptive Filter for Hyperspectral Image

Yao Ding, Weijie Kang, Aitao Yang, Zhili Zhang, Junyang Zhao,
Jie Feng, *Senior Member, IEEE*, Danfeng Hong, *Senior Member, IEEE*, Qinghe Zheng

*Abstract*—Hyperspectral image (HSI) clustering has been a fundamental but challenging task with zero training labels. Currently, some deep graph clustering methods have been successfully explored for HSI due to their outstanding performance in effective spatial structural information encoding. Nevertheless, insufficient structural information utilization, poor feature presentation ability, and weak graph update capability limit their performance. Thus, in this paper, a homophily structure graph learning with an adaptive filter clustering method (AHSGC) for HSI is proposed. Specifically, homogeneous region generation is first developed for HSI processing and constructing the original graph. Afterward, an adaptive filter graph encoder is designed to adaptively capture the high and low frequency features on the graph for subsequence processing. Then, a graph embedding clustering self-training decoder is developed with KL Divergence, with which the pseudo-label is generated for network training. Meanwhile, homophily-enhanced structure learning is introduced to update the graph according to the clustering task, in which the orient correlation estimation is adopted to estimate the node connection, and graph edge sparsification is designed to adjust the edges in the graph dynamically. Finally, a joint network optimization is introduced to achieve network self-training and update the graph. The K-means is adopted to express the latent features. Extensive experiments and repeated comparative analysis have verified that our AHSGC contains high clustering accuracy, low computational complexity, and strong robustness. The code source will be available at https://github.com/DY-HYX.

This work was supported in part by National Key Basic Research Strengthen Foundation of China under Grant 2021-JCJQ-JJ-0871. (Corresponding author: Zhili Zhang, Weijie Kang.)

Z. Zhang, W. Kang, Y. Ding, and A. Yang, J. Zhao are with the key Laboratory of Intelligent Control, PLA Rocket Force University of Engineering, Xi'an 710025, China. (e-mail: 157918018@qq.com; milankang@foxmail.com, dingyao.88@outlook.com; 824360083@qq.com, zhaojy802@sina.com.)

J. Feng are with the Key Laboratory of Intelligent Perception and Image Understanding of Ministry of Education of China, Xidian University, Xi'an 710071, P.R. China (e-mail: jiefeng0109@163.com).

D. Hong is with the Aerospace Information Research Institute, Chinese Academy of Sciences, 100094 Beijing, China, and also with the School of Electronic, Electrical and Communication Engineering, University of Chinese Academy of Sciences, 100049 Beijing, China. (e-mail: hongdf@aircas.ac.cn).

## I. INTRODUCTION

WITH the unique advantage of acquiring continuous spectral information for target objects, hyperspectral image (HSI) has been widely applied in numerous fields [1, 2], e.g., earth observation [3], environmental monitoring [4], military detection [5] and resource exploration [6]. HSI contains rich potential for feature identification and classification, therefore, HSI interpretation has important research significance. However, the high dimensionality, strong redundancy, and high computational complexity contained by HSIs also present significant challenges for their analysis, processing, and interpretation [6, 7]. Traditionally, we often use numerous labeled samples to train interpretation methods and achieve HSI interpretation, which is costly, labor-intensive, and time-consuming [8, 9]. Therefore, the research on unsupervised hyperspectral image classification methods for land-cover classification analysis is of great significance. By grouping similar pixels in HSI into distinct clusters and facilitating the extraction of meaningful information without pre-labeled training samples, Clustering analysis, an unsupervised learning technique, holds promise in overcoming these challenges [10, 11].

Prophase, some clustering methods initially developed for natural images have been adopted for HSI, including *K*-means clustering [12], possibilistic C-means (PCM) [13], spectral clustering [14], fuzzy C-means [15] and density peak clustering [16]. These methods have achieved some success in HSI clustering. At the same time, they only utilize the prototypical features with numerous redundant information, facing challenges in effective dimension reduction, noise handling, and algorithm optimization [17]. Consequently, these limitations result in unideal clustering accuracy, and the widespread applicability has been restricted. After that, to address the redundant information interference, some subspace clustering(SC) algorithms [8] were designed by mapping the high-dimensional image information into a low-dimensional information space, such as parse subspace clustering (SSC) [18], nonnegative matrix factorization (NMF) [19] and low-rank representation (LRR) [20]. However, these methods only focus on extracting spectral features, and the intrinsic relationship of spatially identifiable features is ignored [21]. To flexibly utilize spatial-spectral information, Zhang *et al.* [22] utilized a fuzzy similarity measure to identify the boundaries of different categories. Zhai *et al.* [23] further leveraged spatial information more effectively by adding an $\ell_2$ norm regularization based on SSC. Inspired by graph learning to extract potential relationships between nodes, some graph subspace clustering algorithms were developed. For example, Wang *et al.* [24] proposed a scalable graph-based clustering (SGCNR), in which the computational complexity is reduced by nonnegative relaxation. However, the aforementioned traditional clustering methods are shallow classifiers based on original features, and the handcrafted spectral-spatial features are often relied on, which are often unable to extract representative features [25]. These shortcomings make it difficult to explore deep semantic information and adapt to the spectral variations of HSI.

Inspired by the development of deep clustering learning, various deep clustering methods have been applied to overcome the limitations of traditional clustering methods, in which the hyper-parameters are updated through self-supervised training. The discriminative spatial-spectral features are learned for clustering. We often can divide most of the deep clustering methods into three modules, i.e., deep feature extractor, representative feature expressor, and self-training optimizer. For example, Zhang *et al.* [26] built spatial-spectral similarity graphs based on superpixels, and the spatial-spectral features were simultaneously extracted and decoded by a dual graph autoencoder. Cai *et al.* [17] proposed a contrastive subspace clustering method named NCSC, in which a pooling autoencoder was developed to learn the superpixel-level latent subspace and representation. These methods have made significant attempts to apply deep clustering methods to HSI clustering. At the same time, they are often need more effectively capture intricate spatial structural information, resulting in subpar clustering performances.

Recently, the development of graph clustering methods has made it possible for deep clustering methods to effectively encode spatial structural information, which has transformed deep clustering methods from only processing Euclidean data to non-Euclidean geometric graph structure data. Numerous structural graph clustering networks have been developed, and they have demonstrated remarkable effectiveness in learning structural information through graph embeddings. For example, in [27], a low-pass graph filter is designed to learn smoother structural features, and the clustering accuracy is improved. Luo *et al.* [28] designed a dual transformer-based autoencoder to extract the long-dependency graph features, with which the graph structure features have been precisely presented. To explore target-oriented graph clustering methods, in [29] the graph attention network (GAT) was utilized to aggregate the neighbor node correlation information, and a better clustering performance was guided by developing self-training modules. Meanwhile, to extract higher-order and global structural correlations, some high-order graphs, e.g., pixel-superpixel graph [30], multiview graph, and hypergraph [31], have emerged to capture the structure and spectral-spatial information. Although we have made significant progress in graph-based clustering methods for HSI, the filters in existing graph networks are invariant and need more ability to process signal features for different nodes, resulting in poor clustering accuracy for large-scale HSI [32]. In addition, the mainstream graph networks cannot update the original graph according to different clustering tasks; that is, the graph is fixed and unchanged during the network training, which makes it challenging to correct erroneous edge connections in the original graph, which seriously limits the HSI clustering accuracy improvement with large graph.

To overcome the above questions, we design a novel homophily enhanced structure graph learning with an adaptive filter clustering method (AHSGC) for HSIs. Specifically, a superpixel segmentation method is established to generate a graph structure. Afterward, a graph convolution encoder with an adaptive filter is designed to capture both high-frequency and low-frequency features of the graph adaptively. Furthermore, a graph embedding clustering self-training decoder is designed to obtain appropriate graph representations. Meanwhile, we propose a structure homophily-enhanced learning homophily, in which the orient correlation estimation is introduced to estimate the pairwise correlation between nodes by a hierarchical correlation estimation mechanism. Then, intra-cluster edge recovery and inter-cluster edge removal mechanisms are developed to dynamically adjust the graph structure according to the different clustering tasks. Subsequently, a joint network optimization is introduced, and a self-training loss and the graph reconstruction loss are employed to conduct the graph self-training and update the graph. Finally, we adopt K-means to express the latent features.

The innovative contributions of our AHSGC are as follows:

1) We design joint network optimization to achieve network self-training and update the graph, with which different modules can be integrated into a unit network to improve clustering accuracy.

2) We propose a graph convolution encoder with an adaptive filter to extract both high-frequency and low-frequency components on the graph adaptively.

3) We develop homophily-enhanced structure learning to dynamically adjust the graph structure according to the different clustering tasks.

## II. NOTATIONS AND DEFINITION

*A. Graph and Graph Filter*

*1) Graph definition:*

Let $\mathcal{G} = (\mathcal{V}, \mathcal{E}, \boldsymbol{A})$ be an undirected graph with $N$ nodes, $C$ classes and edge set $\mathcal{E}$, where $\mathcal{V} = \{v_1, v_2, \cdots, v_N\}$, $\boldsymbol{X} \in \mathbb{R}^{N \times D}$ and $\boldsymbol{A} \in \mathbb{R}^{N \times N}$ are the feature matrix and adjacency matrix of the constructed graph, respectively.

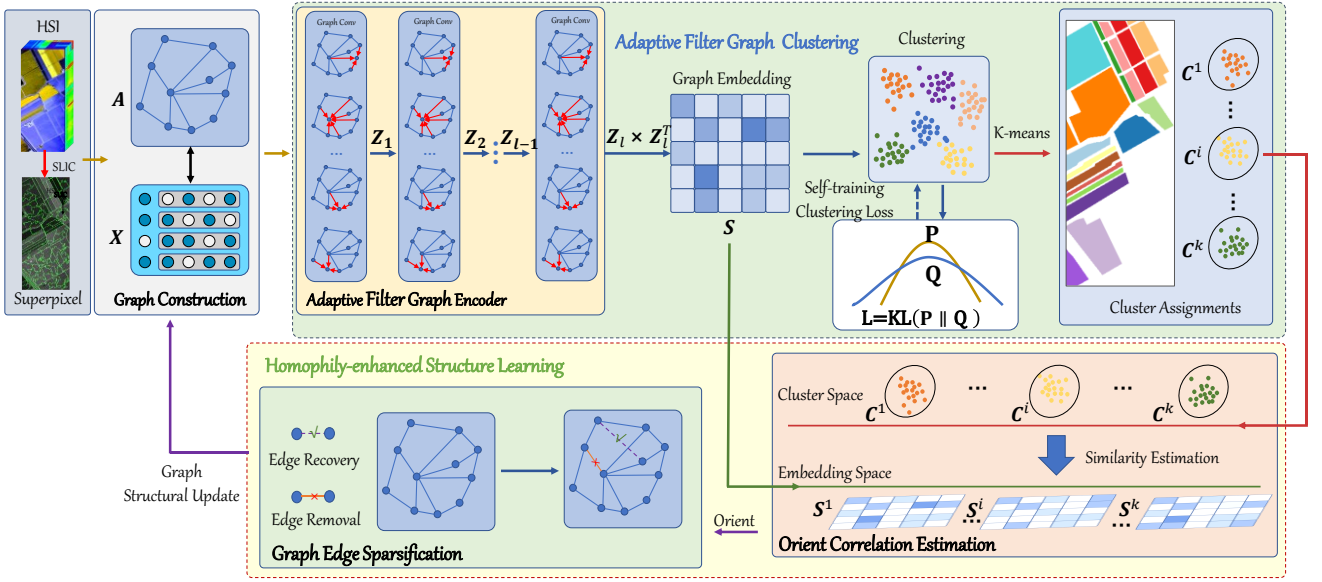

Fig.1. Overview of AHSGC for HSI. AHSGC mainly consists of five parts, i.e., HSI Preprocessing and initial graph construction, adaptive filter graph encoder, graph embedding clustering self-training decoder, homophily-enhanced structure learning, and joint network optimization module. The HSI Preprocessing and initial graph construction are proposed to transform the pixels in HSI to locality-preserving regions and construct the original graph. An adaptive filter graph encoder is designed to extract and filter the low and high frequency features on the graph. The graph embedding clustering self-training decoder is used to conduct the self-training and decoder the extracted features. The homophily-enhanced structure learning is developed to update the graph according to the clustering task. The joint network optimization module is proposed to enhance the structural consistency and graph edge sparsification of the graph and train the proposed network.

We summery the notions in Table I.

**TABLE I**
**MAIN NOTIONS SUMMERY**

| Notation | Meaning |
|---|---|
| $\mathcal{G}=(\mathcal{V},\mathcal{E},\mathbf{A})$ | Graph $\mathcal{G}$ with edge $\mathcal{E}$ and node $V$ |
| $N$ | Graph node number |
| $\mathbf{X} \in \mathbb{R}^{N \times D}$ | Node feature matrix |
| $\mathbf{X}_i$ | $i$-th node feature |
| $\mathbf{A} \in \mathbb{R}^{N \times N}$ | Adjacency matrix |
| $S_i$ | The $i$th superpixel |
| $n_i$ | The number of pixels contained in $S_i$ |
| $\mathbf{Z} \in \mathbb{R}^{hw \times N}$ | The correlation matrix |
| h, w, and b | The height, width, and number of bands of the HSI |
| $\mathbf{S} \in \mathbb{R}^{N \times N}$ | Node similarity matrix |
| $\mathbf{H}$ | Graph Laplacian filter |
| $\mathbf{F}_A$ | Adaptive graph filter |
| $\mathbf{A}_{rw}$ | Randomly wandering normalized affine matrix |
| $\mathbf{Z}_i$ and $\mathbf{Z}$ | The $i$-th layer and graph encoder output |
| $\mathbf{Z} \in \mathbb{R}^{N \times D}$ | The clustering-oriented node features |
| $\xi$ | Intra-cluster edge recovery ratio |
| $\eta$ | Inter-cluster edge removal ratio |
| $\gamma$ | high-confidence node extraction ratio |
| $\mathcal{L}_c, \mathcal{L}_g$ | Self-traning loss and graph reconstruction loss |

*2) Graph filter definition:*

For a discrete signal $\hat{f}_{in}(\xi)$, after the Fourier convolution operation [33], the output time-domain signal $f_{out}(t)$ can be expressed as

$$f_{out}(t) = \int_{-\infty}^{+\infty} \hat{f}_{in}(\xi)\hat{h}(\xi)e^{2\pi i \xi t}d\xi$$
$$= \int_{-\infty}^{+\infty} f_{in}(\tau)h(t-\tau)d\tau \quad (1)$$
$$= (f_{in} * h)(t)$$

where the unit of $\xi$ is $Hz$, and $\hat{h}(\xi)$ is a frequency response function.

According to Eq. (1), the graph convolution operation on discrete graph node signal $\hat{h}(\lambda_k)$ can be defined as

$$f_{out}(i) = \sum_{k=1}^{n} \hat{f}_{in}(\lambda_k)\hat{h}(\lambda_k)\mathbf{v}_k(i) \quad (2)$$

where $\hat{f}_{in}(\lambda_k)$ represents the strength of the graph signal at frequency $\lambda_k$, $\mathbf{v}_k(\cdot)$ is the Fourier basis of the graph signal, $i$ represents the $i$th element of $\mathbf{v}_k(\cdot)$. In Eq. (2), the integral operation in Eq. (1) is replaced with the sum operation.

Let $\hat{f}_{in} = U^T f_{in}$, we can simplify Eq. (2) to matrix form as

$$f_{out} = \sum_{k=1}^{n} h(\lambda_k)\hat{f}_{in}(\lambda_k)\mathbf{v}_k = \begin{bmatrix} | & | & \cdots & | \\ \mathbf{v}_1 & \mathbf{v}_2 & \cdots & \mathbf{v}_n \\ | & | & \cdots & | \end{bmatrix} \begin{bmatrix} h(\lambda_1)\hat{f}_{in}(\lambda_1) \\ h(\lambda_2)\hat{f}_{in}(\lambda_2) \\ \cdots \\ h(\lambda_n)\hat{f}_{in}(\lambda_n) \end{bmatrix}$$
$$= \begin{bmatrix} | & | & \cdots & | \\ \mathbf{v}_1 & \mathbf{v}_2 & \cdots & \mathbf{v}_n \\ | & | & \cdots & | \end{bmatrix} \begin{bmatrix} h(\lambda_1) & \cdots & 0 \\ \vdots & \ddots & \vdots \\ 0 & \cdots & h(\lambda_n) \end{bmatrix} \begin{bmatrix} - & \mathbf{v}_1^T & - \\ - & \mathbf{v}_2^T & - \\ & \cdots & \\ - & \mathbf{v}_n^T & - \end{bmatrix} f_{in} \quad (3)$$
$$= U \begin{bmatrix} h(\lambda_1) & \cdots & 0 \\ \vdots & \ddots & \vdots \\ 0 & \cdots & h(\lambda_n) \end{bmatrix} U^T f_{in} = U\Lambda_h U^T f_{in} = \mathbf{H}f_{in}$$

Accordingly, the graph filter can be defined as

$$\boldsymbol{H} = \boldsymbol{U}\boldsymbol{\Lambda}_h\boldsymbol{U}^T \in \mathbb{R}^{n\times n}, \boldsymbol{H}: \mathbb{R}^n \to \mathbb{R}^n \quad (4)$$

where $\boldsymbol{\Lambda}_h$ is the frequency response matrix of the graph filter.

*B. Homophily-enhanced Graph Learning*

Given a graph $\mathcal{G} = \{X, A\}$ with $C$ classes, we assign the node $v_i$ as $y_i$. The edge homogeneity concept is the ratio of edges in the same class to the total number of edges in the graph, it can be defined as

$$h(\mathcal{G}, \{y_i; v_i \in \mathcal{V}\}) = \frac{1}{|\mathcal{E}|}\sum_{(v_i,v_j)\in\mathcal{E}} \mathbb{I}(y_i = y_j) \quad (5)$$

where $\mathbb{I}(\cdot)$ is an indicator function

The purpose of graph clustering is to classify node set $\mathcal{V}$ into class $C$, that is, to assign node $v_i$ into specific cluster with its specific features, which can be expressed as

$$\begin{cases} \min \sum_{i=1}^{m}\sum_{j=1}^{m} d(i,j) = \min \sum_{i=1}^{m}\sum_{j=1}^{m} \|f(x_i) - f(x_j)\|_2^2 \\ \max \sum_{\alpha=1}^{c}\sum_{\beta=1}^{c} d(\alpha,\beta) = \max \sum_{\alpha=1}^{c}\sum_{\beta=1}^{c} \|y(c_\alpha) - y(c_\beta)\|_2^2 \end{cases} \quad (6)$$

where $f(\bullet)$ and $y(\bullet)$ are functions to map the node's optimal spatial-spectral information and calculate the feature of the cluster, respectively. $m$ is the node number within the specific cluster.

III. PROPOSED METHODOLOGY

In this section, the overall framework of AHSGC is elaborated first. Afterward, the structure details of AHSGC, i.e., Adaptive Filter Graph Encoder, Graph Embedding Clustering Self-Training Decoder, Homophily-enhanced Structure Learning and Joint Network Optimization are fully introduced. Finally, HSI Preprocessing and Initial Graph Construction are introduced.

*A. Overall AHSGC Framework*

The proposed AHSGC mainly consists of five modules, i.e., Adaptive Filter Graph Encoder, Graph Embedding Clustering Self-Training Decoder, Homophily-enhanced Structure Learning, Joint Network Optimization, and HSI Preprocessing and Initial Graph Construction module, the overall framework of AHSGC is shown in Fig.1. The main functions of the five modules are as follows

● **Adaptive Filter Graph Encoder:** The superpixel-level graph is the input of AHSGC. To extract and filter the low and high-frequency features on the graph for clustering, an adaptive graph filter is designed by conducting neighbor information aggregation with an adaptive graph filter.

● **Graph Embedding Clustering Self-Training Decoder:** an auxiliary distribution is adopted to conduct the self-training for AHSGC, and then the k-means is utilized to implement clustering.

● **Homophily-enhanced Structure Learning:** To obtain sparser edges, we developed inter-cluster edge removal and intra-cluster edge recovery, with which the graph structure can be dynamically adjusted according to the different clustering tasks.

● **Joint Network Optimization:** A novel joint objective loss is proposed to enhance the structural consistency and graph edge sparsification of the graph and train the proposed network. Finally, the K-means is adopted to express the latent features.

● **HSI Preprocessing and Initial Graph Construction:** A superpixel segmentation method is established to transform the HSIs from pixels to regions while preserving the local spatial-spectral structure information and reducing the node number.

In our AHSGC, five clustering modules are joined in an end-to-end network for HSI clustering, and each portion interacts with the other.

*B. Adaptive Filter Graph Encoder*

Typically, the low-pass filters can obtain low-frequency information on the graph, filtering out high-frequency information, and the smoother graph node signal features are obtained. While the high-pass filters have the opposite function (show in Fig.2). Importantly, the graph node signals contain low-pass signals and high pass signals, both of which have significant impacts on downstream clustering tasks [34]. How to comprehensively utilize the two types of feature signals is of great significance for improving clustering accuracy.

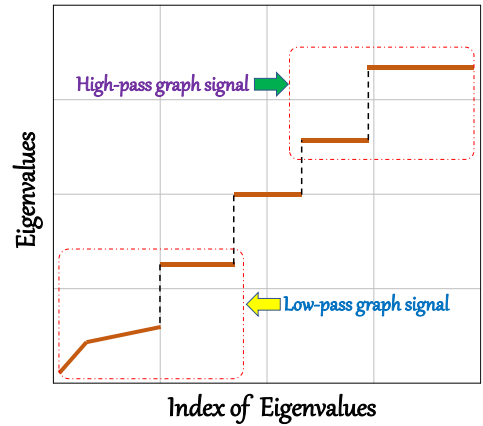

Fig.2. Illustration of the role of low-pass and high-pass filters.

In general, graph filters are associated with Laplacian matrices, affinity matrices, and their variants, respectively. The low-pass filters $\boldsymbol{H}_l$ and high-pass filters $\boldsymbol{H}_h$ can be expressed as

$$\boldsymbol{H}_l = \boldsymbol{A}_{rw}\boldsymbol{X}, \boldsymbol{H}_h = \boldsymbol{L}_{rw}\boldsymbol{X} \quad (7)$$

where $\boldsymbol{A}_{rw}$ is the randomly wandering normalized affine matrix, expressed as $\boldsymbol{A}_{rw} = \boldsymbol{D}^{-1}\boldsymbol{A}$, $\boldsymbol{L}_{rw}$ is the corresponding Laplacian matrix, namely $\boldsymbol{L}_{rw} = \boldsymbol{I} - \boldsymbol{A}_{rw}$.

According to the introduction in Section II-A, given a convolution kernel $f$ and a graph signal $x$, the convolution operation can be expressed as follows

$$f * x = \boldsymbol{U}\big((\boldsymbol{U}^T f) \odot (\boldsymbol{U}^T x)\big) = \boldsymbol{U}g^\theta \boldsymbol{U}^T x \quad (8)$$

where $g^\theta = \boldsymbol{U}^T f$ denotes the representation of $f$ in the spectral domain, ⊙ is the matrix dot multiplication.

If $A_{rw}$ and $L_{rw}$ are treated as convolution kernels as Eq.(7), then we can filter the graph signal $x$ as

$$A_{rw} * x = U(g_l^\theta)U^T x, L_{rw} * x = U(g_h^\theta)U^T x \quad (9)$$

where $g_l^\theta$ and $g_h^\theta$ are the representation of $A_{rw}$ and $L_{rw}$ in the spectral domain. If $g_l^\theta = I - \Lambda$ and $g_h^\theta = \Lambda$, the Eq.(9) is rewrite as

$$U(g_l^\theta)U^T x = \sum_i (1-\lambda_i) u_i u_i^T x$$
$$U(g_h^\theta)U^T x = \sum_i \lambda_i u_i u_i^T x \quad (10)$$

In Eq.(10), if the eigenvector $\lambda$ is set as $\lambda > 1$, the difference between $x_i$ is amplified in the projection $\sum_i \lambda_i u_i u_i^T x$, then the non-smooth signals are obtained. if $\lambda<1$, the difference between $x_i$ is reduced in the projection $\sum_i \lambda_i u_i u_i^T x$, and the smooth signals are obtained. After analysis, we note that low-pass filters can capture similarity information, while high-pass filters are the opposite. Therefore, both types of filters play essential roles in graph networks. To capture the complete node information, an adaptive graph filter $F_A$ is designed, namely

$$F_A = \mu \cdot (S_{rw})^k X + (1-\mu) \cdot (I - S_{rw})^k X \quad (11)$$

where $X$ represents the input spectral features, $\mu > 0$ denotes learnable parameter to balance the high-pass and low-pass information. $S_{rw}$ is the normalized Laplacian matrices associated to $A_{rw}$.

To capture the high-pass and low-pass presentations of the constructive graph, we stack $t$ layers of graph Laplacian filters as follows

$$X_s = \left(\prod_{i=1}^{t} F_A\right) X = F_A^t X \quad (12)$$

where $X$ represents the input graph node spectral features, $F_A^t$ is the stacked $t$ layers of adaptive graph filters, and $X_s$ is graph node presentations after $t$ layers convolution.

C. Graph Embedding Clustering Self-Training Decoder

*1) Graph Embedding:* The adaptive filter graph encoder module extracts high-level structural-semantic information, while the high-pass and low-pass presentations are learned. To achieve an efficient fusion representation of structural information and spectral features, a linear transformation mechanism is developed to encode the graph structure as follows

$$Z = \left(\prod_{i=1}^{t} F_A\right) Z_0 W = F_A^t Z_0 W \quad (13)$$

where $Z_i$ and $Z$ are the $i$-th layer and graph encoder output, respectively, $Z_0 = X$. $W$ denotes the linear transformation weight matrix.

*2) Self-Training Clustering:* As analyzed above, we can learn a high-pass and low-pass graph presentations with an adaptive filter. However, the high-level structural-semantic information cannot be used for clustering, and unlabeled is also a barrier to graph clustering. To address the problems, a self-training clustering mechanism is designed by using an auxiliary distribution. Concretely, we adopt a Student's t-distribution to measure the similarity between the $k$-th cluster center $\mu_k$ and the node embedding $z_i$ as follows

$$q_{ik} = \frac{(1 + \|z_i - \mu_k\|^2)^{-1}}{\sum_k^K (1 + \|z_i - \mu_k\|^2)^{-1}} \quad (14)$$

where $q_{ik}$ is clustering assignment distribution, which can be considered as the soft assignment of node $v_i$ to $\mu_k$. Then we can optimize the clustering distribution by minimizing the KL divergence, namely

$$\mathcal{L}_c = KL(P \| Q) = \sum_i \sum_k p_{ik} \log \frac{p_{ik}}{q_{ik}} \quad (15)$$

$$p_{ik} = \frac{q_{ik}^2 / \sum_j q_{jk}}{\sum_k^K (q_{ik}^2 / \sum_j q_{jk})} \quad (16)$$

where $p_{ik}$ is the auxiliary distribution, as described in Eq.(16), the $P$ raises $Q$ to the second power, with which the $Q$ would be a sharp distribution, and it can avoided the clustering collapsing into a single cluster [35].

D. Homophily-enhanced Structure Learning

*1) Orient Correlation Estimation:* Given an intermediate clustering result, generally, the inner product is used to measure the pairwise correlation between nodes. While, in practice, we have found that this is a suboptimal approach, because the clustering results contain considerable noise during the network training [36]. To address this problem, a hierarchical correlation estimation mechanism is designed. Precisely, the node similarity is estimated incorporating guidance from both the latent space in a hierarchical manner and clustering. First, the hard pseudo-label is calculated

$$c_i = \arg\max_k q_{ik} \quad (17)$$

According to Eq. (17), all nodes' pseudo-labels are obtained, then, the nodes' top $\gamma$ percentage is selected in each cluster $k$ with the highest assignment probability. Thus, a node subset is obtained, namely

$$\Gamma^k = \{v_i | c_i = k, Rank|_{\{v_i | c_i = k\}}(q_{ik}) \leq \gamma * |\{v_i \mid c_i = k\}|\} \quad (18)$$

where $\Gamma^k$ is the top-confident node subset, $Rank_{\{v_i|c_i=k\}}(q_{ik})$ denotes the cluster assignment ranking probability of node $v_i$ in cluster $k$. With Eq. (18), we can reduce the noise effects on node clustering. However, if the edge weight is still determined by cluster space, confident but incorrect segments will still undermine the similarity matrix accuracy. To overcome the limitations, in this paper, we adopt the latent space to estimate the node similarity rather than the clustering space.

In Eq.(13), the node embedding matrix $Z$ is calculated, we can estimate the similarity between all nodes by dot product

$$S = ZZ^T \quad (19)$$

where $S$ is node similarity. Different from Eq.(19), in this paper, we calculate $S$ within $\Gamma^k$, and the corresponding node embedding node similarity $S_{ij}^k$ is expressed as

$$S_{ij}^k = Z_i^k Z_j^{k^T} \in \mathbb{R}^{|\Gamma^k| \times |\Gamma^k|} \quad (20)$$

where $Z_i^k$ is the node embedding matrix of $i$-th class, $|\Gamma^k|$ denotes the number of nodes in $\Gamma^k$.

Eq.(20) comprehensively utilizes the embedding similarity matrix and soft assignments to identify the incorrect or missing edges, with which the edge modification accuracy is significantly enhanced. Therefore, a more reliable and precise graph structure learning is achieved.

*2) Graph Edge Sparsification:* Most existing methods keep the graph unchanged during the network training, which means that the edges in the graph remain unchanged. This strategy has two drawbacks, e.g., incorrect edge connections in the graph cannot be corrected, and it brings unnecessary difficulties to clustering. Inspired by Eq.(6), we propose a graph edge sparsification, including inter-cluster edge removal and intra-cluster edge recovery, to make the nodes within the cluster more closely connected and the nodes between clusters sparser (show as Fig.3).

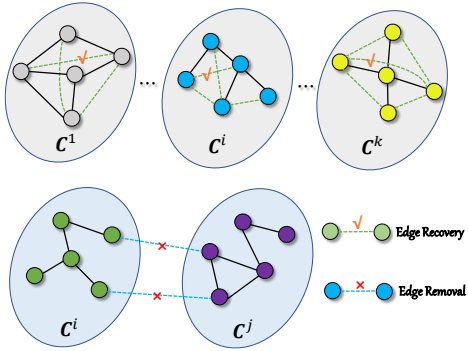

Fig.3. The mechanism of intra-cluster edge recovery and Inter-cluster edge removal.

**Intra-cluster edge recovery:** To increase connections within the cluster, the edges are added to the top $\xi$ percent of node pairs containing the largest similarity $S_{ij}^k$ within cluster $k$, namely

$$\mathcal{E}_{rc}^k = \left\{(v_i, v_j) | \text{Rank}(S_{ij}^k) \leq \xi * |\mathcal{E}| * \frac{N_k}{N}, \text{ and } c_i = c_j = k \right\} \quad (21)$$

where $N_k$ denotes the node number in $\Gamma^k$, $\mathcal{E}$ is the edge-set on the current graph learning, $\text{Rank}(S_{ij}^k)$ indicates the node pair $(v_i, v_j)$ similarity is ranked in order within $S^k$. We should note that existing edges are unconsidered in the ranking process. We can express the complete intra-cluster edge recovery set $\mathcal{E}_{rc}$ as

$$\mathcal{E}_{rc} = \bigcup_{k}^{K} \mathcal{E}_{rc}^k \quad (22)$$

**Inter-cluster edge removal**: This strategy aims to remove unnecessary edge connections between clusters, weaken inter cluster connections, and facilitate clustering. The expression is as follows

$$\mathcal{E}_{rm} = \{(v_i, v_j) | \text{Rank}(S_{ij}) \geq (1-\eta) * |\mathcal{E}|, (v_i, v_j) \in \mathcal{E}, c_i \neq c_j\} \quad (23)$$

where $\mathcal{E}$ is the edge-set on the current graph learning. $\eta$ is the percentage of retained edges.

*3) Homophily-enhanced Structure Learning:* We have demonstrated the recovery and removal edge set in Eq.(22) and Eq.(23). The edge set during the graph learning can be expressed as $\bar{\mathcal{E}} = \mathcal{E} - \mathcal{E}_{rm} + \mathcal{E}_{rc}$. The adjacency matrix $\bar{A}$ can be computed as

$$\bar{A} = A - A_{\mathcal{E}_{rm}} + A_{\mathcal{E}_{rc}} \quad (24)$$

where $A_{\mathcal{E}_{rm}}$ and $A_{\mathcal{E}_{rc}}$ are the adjacency matrices of $\mathcal{E}_{rm}$ and $\mathcal{E}_{rc}$. Then, we reconstruct the graph by updating the adjacency matrix, namely

$$\mathcal{L}_g = \|ZZ^T - \bar{A}\|_F^2 \quad (25)$$

where $\mathcal{L}_g$ is the graph reconstruction loss.

*E. Joint Network Optimization*

We show the self-training clustering loss in Eq.(15), and the structure graph reconstruction loss is demonstrated in Eq.(25), then the overall objective $\mathcal{L}_O$ of AHSGC can be formulated as

$$\mathcal{L}_O = \mathcal{L}_c + \mathcal{L}_g \quad (26)$$

In AHSGC, the $p_{ik}$ is treated as "ground-truth", and the hyper-paraments are updated $T$ iteration by minimizing the overall objective $\mathcal{L}_O$. The clustering result for node $v_i$ is formulated as

$$Y_i = \underset{k}{\text{argmax}}\, q_{ik} \quad (27)$$

Therefore, the adaptive filter graph encoder is used to capture the latent features of the graph, and the graph embeddings $Z$ is derived. Subsequently, the $K$-means is adapted to perform clustering on $Z$ based on the initialized cluster centers $\mu_k$. The learning process is detailed in Algorithm 1.

| Algorithm 1: AHSGC. | |
|---|---|
| | **Input**: The input graph; Initial centroids (Cluster) number $k$; Iterations number $T$; Number of the adaptive graph filter layer $t$. |
| | **Output**: Clustering results $Q$. |
| 2: | Initialization: $t$; $Z^0 = X$; |
| 3: | Generate the homogeneous region and construct the superpixel-based graph; |
| 4: | **for** $t$=1 to $T$ **do** |
| 5: | Apply $t$-layers graph convolution with adaptive graph filters on $Z^0$ to obtain the latent $Z$ by Eq.(13); |
| 6: | The self-training clustering loss $\mathcal{L}_c$ is formulated as by Eq.(15); |
| 7: | The homophily-enhanced structure learning is implemented by Eq.(21) and Eq.(23); |
| 8: | The adjacency matrix $\bar{A}$ is calculated by Eq.(24); |
| 9: | The graph reconstruction loss is calculated by Eq.(25) |
| 10: | The weight matrices are updated by minimizing $\mathcal{L}_O$ with Adam optimizer. |
| 11: | **end** |
| 12: | The clustering results $Q$ is obtained by applying $K$-means on $Z$. |

*F. HSI Preprocessing and Initial Graph Construction*

In our paper, the simple linear iterative cluster (SLIC) [37] method is first applied to divide the entire HSI into many spatially connected superpixels. Furthermore, the superpixel-level node feature $X$ is calculated as the average spectral features of the pixels contained in that superpixel. The $A \in \mathbb{R}^{N \times N}$ is formulated as

$$A_{ij} = \begin{cases} e^{-\rho \|X_i - X_j\|^2}, & \text{if } X_i \in N_t(X_j) \text{ or } X_j \in N_t(X_i) \\ 0, & \text{otherwise} \end{cases} \quad (28)$$

where $X_i$ denotes $i$-th superpixel-level node feature, $\rho = 0.2$, $N_t(X_i)$ presents the node $v_i's$ $t$-hop neighbors.

To explore the connection between pixel and superpixel, image backprojection operations are developed to transform data features from pixel to superpixel, it can be represented as

$$Q_{i,j} = \begin{cases} 1, & \text{if } x_i \in S_i \\ 0, & \text{otherwise} \end{cases} \quad (29)$$

where $Q \in \mathbb{R}^{hw \times N}$ is the correlation matrix to record all pixel contained in HSI locating in which superpixel.

Graph projection can encode the original pixel level HSI into superpixel-level graph node features through matrix multiplication, which is formulated as

$$V = \text{Projection}(X; Q) = \hat{Q}^T \text{Flatten}(X) \quad (30)$$

where $\hat{Q}$ is $Q$ normalized by column, i.e., $\hat{Q}_{i,j} = Q_{i,j} / \sum_m Q_{m,j}$. Flatten(·) represents flattening HSI according to spatial dimensions.

## IV. EXPLEMENTS

In this section, extensive experiments and analyses are introduced to assess the clustering performance of the proposed AHSGC. Specifically, the three well-known HSI datasets and experimental setups are introduced (Section IV-A and Section IV-B); subsequently, the quantitative and visual HSI clustering performances of AHSGC are compared with nine state-of-the-art algorithms (Section IV-C and Section IV-D); furthermore, the runtime and model complexity are compared with investigated methods (Section IV-E); in addition, the influence of hyperparameters are analyzed (Section IV-F); moreover, the clustering visualizations are demonstrated (Section IV-G), finally, we implement some ablation experiments to evaluate the rationality of the proposed AHSGC design (Section IV-H).

### A. Data Description

In this paper, we validate our method using the Salina (SA), Trento, and Pavia University (PU) datasets. The SA dataset contains a dimension of 512×217 and 204 bands, totaling 16 land cover classes. The Trento dataset has an image size of 600×166, with 63 spectral channels ranging from 0.40 to 0.98 μm, covering 6 land cover classes. The PU dataset consists of an image composed of 103 spectral bands with dimensions of 610×340, encompassing 9 land cover classes. We show the specific land covers and corresponding quantities of the three datasets in Table II, and the corresponding ground-truth is shown in Fig. 4.

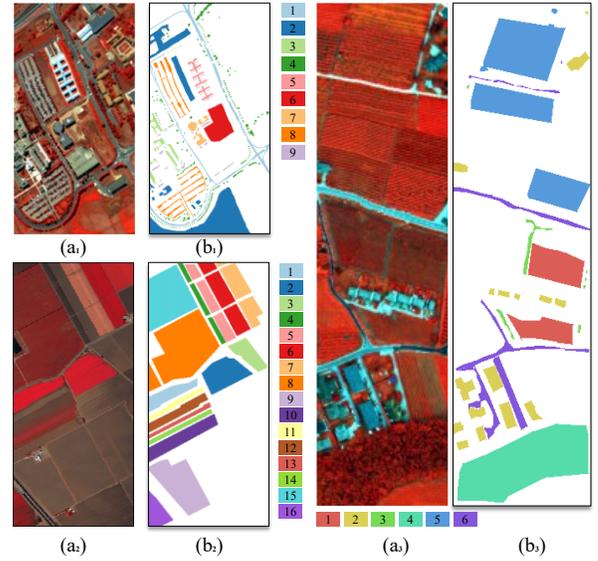

Fig.4. Three dataset details. (a₁), (a₂), and (a₃) are the false-color maps of SA, PU, and Trento datasets. (b₁), (b₂), and (b₃) are the ground-truth maps.

### B. Experimental Setup

*1) Experiment Settings:* The overall classification accuracy (OA), kappa coefficient (κ), normalized mutual information (NMI), adjusted rand index (ARI), and Purity are adopted as quantitative evaluation metrics. Among them, OA is used to record the overall clustering accuracy; κ is an indicator adopted to evaluate the consistency degree; NMI denotes the similarity of clustering results; ARI represents the degree of agreement between clustering results and the ground truth, Purity∈ [0, 1] is a simple and transparent evaluation metric. In our AHSGC, seven main parameters, i.e., graph number $N$, encoder layer number $l$, iterations number $T$, learning rate $L$, intra-cluster edge recovery ratio $\xi$, inter-cluster edge removal ratio $\eta$, and high-confidence node extraction ratio $\gamma$ should be preset. Their optimal values are shown in Table III and will be analyzed in Section IV-F.

TABLE II
THE DETAILS OF EACH LAND-COVER CLASS OF THREE DATASETS USED IN THE PAPER

| NO. | SA | | PU | | Trento | |
|---|---|---|---|---|---|---|
| | Name | Pixels | Name | Pixels | Name | Pixels |
| 1 | Weed 1 | 2009 | Asphalt | 6631 | Apple trees | 4034 |
| 2 | Weed 2 | 3726 | Meadows | 18649 | Buildings | 2903 |
| 3 | Fallow | 1976 | Gravel | 2099 | Ground | 479 |
| 4 | Fallow-plow | 1394 | Trees | 3064 | Woods | 9123 |
| 5 | Fallow-smooth | 2678 | Painted metal sheets | 1345 | Vineyard | 10501 |
| 6 | Stubble | 3959 | Bare Soil | 5029 | Roads | 3174 |
| 7 | Celery | 3579 | Bitumen | 1330 | | |
| 8 | Grapes untrained | 11271 | Self-Blocking Bricks | 3682 | | |
| 9 | Soil | 6203 | Shadows | 947 | | |
| 10 | Corn | 3278 | | | | |
| 11 | Lettuce-4wk | 1068 | | | | |
| 12 | Lettuce-5wk | 1927 | | | | |
| 13 | Lettuce-6wk | 916 | | | | |
| 14 | Lettuce-7wk | 1070 | | | | |
| 15 | Vineyard-untrained | 7268 | | | | |
| 16 | Vineyard-trellis | 1807 | | | | |
| Total | | 54129 | | 42776 | | 30214 |

TABLE III
PRESET OPTIMAL PRAMENTERS IN AHSGC

| Dataset | $N$ | $l$ | $T$ | $L$ | $\gamma$ | $\xi$ | $\eta$ |
|---------|-----|-----|-----|------|----------|-------|--------|
| Salinas | 580 | 5 | 50 | 5e-4 | 0.3 | 0.5 | 0.05 |
| PU | 800 | 5 | 50 | 5e-4 | 0.5 | 0.5 | 0.05 |
| Trento | 550 | 5 | 50 | 5e-4 | 0.3 | 0.5 | 0.05 |

*2) Compare Baseline:* To comprehensively analyze the effectiveness of the proposed method, we compares AHSGC with nine selected baselines, e.g., K-means, Fuzzy C-means (FCM), Possibilistic C-means (PCM), NCSC [17], DFCN [38], SDCN [39], EGAE [40], AdaGAE [41], and DAEGC [29]. Among them, K-means, FCM, and PCM are traditional clustering methods, NCSC and DAEGC are subspace clustering and graph clustering methods, DFCN, SDCN, EGAE, AdaGAE, and DAEGC are deep clustering methods, and DAEGC, EGAE and AdaGAE are graph autoencoder methods. The hyper-parameters and network structures of these methods follow the suggestions in the original papers.

*3) Implementation details:* In this paper, all investigated experiments are conducted on NVIDIA Titan RTX using the Pytorch framework. For a fair comparison, all methods are executed ten times for the sake of eliminating any bias caused by the random selection of training samples.

*C. Quantitative and Qualitative Performances Comparison*

In this section, we compare AHSGC with nine other clustering methods to demonstrate its clustering performance through quantitative and qualitative performance analyses. In addition, to be more intuitive, the best performances and the second-best performances are marked in bold and underlined.

*1) Quantitative analysis of the SA dataset:* The different clustering methods' performances on SA are shown in Table IV. From Table IV, we can note that our AHSGC achieves the best clustering performance with 83.60%, 81.62%, 77.42%, and 83.68% in terms of OA, Kappa, NMI, ARI, and Purity, respectively, which is improved by 4.47%, 5.60%, 0.73%, 5.59%, and 0.11% compared with the second-best results. Moreover, NCSC, EGAE, and AdaGAE achieve relatively good clustering results, which shows that these methods have good adaptability to noise interference and class imbalance sample clustering. Moreover, due to the remarkable ability of graph-based methods to exploit the interrelationships between nodes, DAEGC, EGAE, and AdaGAE achieve relatively good clustering results. Due to lacking the ability to capture high-lever features of HSI, the traditional methods, including k-mean, FCM, and PCM) undoubtedly achieve poor results. Specifically, the OAs of the three methods are 67.99%, 56.73%, and 55.63%, respectively. Furthermore, compared with other graph-based method, an adaptive filter can adaptively extract the high-frequency and low-frequency information for subsequent HSI clustering tasks, improving the performance of AHSGC.

TABLE IV
QUANTITATIVE EXPERIMENTAL CLASSIFICATION RESULTS ON SALINAS

| No. | k-means | FCM | PCM | NCSC | DFCN | SDCN | EGAE | AdaGAE | DAEGC | AHSGC |
|-----|---------|-----|-----|------|------|------|------|--------|-------|-------|
| 1 | 0.9985 | 0.4179 | 0.6456 | 0.0000 | **1.0000** | 0.0000 | **1.0000** | 0.0000 | **1.0000** | **1.0000** |
| 2 | 0.5698 | 0.5524 | 0.8125 | **1.0000** | 0.9990 | **1.0000** | **1.0000** | 0.9997 | 0.9997 | **1.0000** |
| 3 | 0.9741 | 0.8501 | 0.0000 | 0.6822 | **1.0000** | 0.3426 | **1.0000** | **1.0000** | 0.0000 | 0.9413 |
| 4 | **0.9865** | 0.0000 | 0.0000 | 0.0000 | 0.0000 | 0.0000 | 0.0000 | 0.0000 | 0.0000 | 0.0000 |
| 5 | 0.7856 | 0.5540 | 0.9628 | **0.9988** | 0.7308 | 0.9816 | 0.0000 | 0.9572 | 0.6502 | 0.9709 |
| 6 | 0.9952 | 0.9990 | 0.9906 | 0.9941 | 0.9986 | 0.9530 | 0.9886 | 0.9900 | **1.0000** | 0.9914 |
| 7 | 0.4810 | 0.0000 | 0.9943 | **1.0000** | 0.9561 | 0.0000 | 0.9983 | 0.9999 | 0.9718 | 0.9922 |
| 8 | 0.7309 | 0.7772 | 0.3875 | 0.8026 | 0.4320 | 0.7633 | 0.5131 | 0.5631 | 0.6683 | **0.9403** |
| 9 | 0.9750 | 0.6969 | 0.5829 | 0.9076 | 0.9063 | 0.8643 | 0.8326 | 0.9832 | 0.9141 | **0.9969** |
| 10 | 0.6169 | 0.5507 | 0.4831 | **0.9900** | 0.7215 | 0.5519 | 0.8962 | 0.9000 | 0.2379 | 0.9356 |
| 11 | 0.3370 | 0.0000 | 0.7462 | **0.9531** | 0.5511 | 0.0000 | 0.0000 | 0.8702 | 0.0000 | 0.0000 |
| 12 | 0.0273 | 0.0047 | 0.8357 | 0.0000 | 0.5051 | 0.2287 | **1.0000** | 0.4926 | 0.8583 | 0.1012 |
| 13 | 0.3548 | 0.8134 | **0.9582** | 0.0000 | 0.0656 | 0.0690 | 0.0000 | 0.0000 | 0.0000 | 0.0000 |
| 14 | 0.5584 | 0.6512 | 0.1390 | 0.0000 | 0.0013 | **0.7632** | 0.0000 | 0.0000 | 0.0000 | 0.0000 |
| 15 | 0.4994 | 0.4279 | 0.2570 | 0.8563 | 0.9477 | 0.0980 | **0.9692** | 0.9436 | 0.8176 | 0.7881 |
| 16 | 0.0000 | 0.8321 | 0.4382 | **1.0000** | 0.2911 | 0.0000 | **1.0000** | 0.9575 | 0.9723 | **1.0000** |
| OA (%) | 0.6799 | 0.5673 | 0.5563 | 0.7881 | 0.7193 | 0.5382 | 0.7361 | 0.7683 | 0.7066 | **0.8360** |
| Kappa | 0.6572 | 0.5431 | 0.5472 | 0.7602 | 0.6900 | 0.4793 | 0.7106 | 0.7382 | 0.6741 | **0.8162** |
| NMI | 0.7356 | 0.6859 | 0.6631 | 0.8514 | 0.8288 | 0.6589 | 0.8769 | 0.8251 | 0.7841 | **0.8587** |
| ARI | 0.5466 | 0.4693 | 0.4539 | 0.7068 | 0.6278 | 0.4831 | 0.7183 | 0.6532 | 0.6096 | **0.7742** |
| Purity | 0.7132 | 0.7016 | 0.6708 | 0.7893 | 0.7324 | 0.5765 | 0.8357 | 0.8002 | 0.7129 | **0.8368** |

TABLE V
QUANTITATIVE EXPERIMENTAL CLASSIFICATION RESULTS ON PU

| No. | k-means | FCM | PCM | NCSC | DFCN | SDCN | EGAE | AdaGAE | DAEGC | AHSGC |
|---|---|---|---|---|---|---|---|---|---|---|
| 1 | 0.8234 | 0.5732 | 0.5538 | 0.4238 | 0.4051 | 0.6234 | 0.3029 | **0.8521** | 0.3759 | 0.4416 |
| 2 | 0.3581 | 0.3265 | 0.4633 | 0.5933 | 0.4008 | 0.3803 | 0.5012 | 0.5329 | **0.7935** | 0.7466 |
| 3 | 0.2167 | 0.0024 | 0.6027 | 0.6902 | 0.7038 | 0.1687 | **0.9900** | 0.7194 | 0.9609 | 0.4383 |
| 4 | 0.5943 | 0.5367 | 0.1648 | 0.0270 | 0.3547 | 0.4025 | 0.2744 | **0.7237** | 0.6505 | 0.6364 |
| 5 | 0.6620 | 0.7124 | 0.0000 | 0.9735 | 0.6892 | 0.6610 | **1.0000** | 0.0000 | **1.0000** | **1.0000** |
| 6 | 0.4255 | 0.3529 | 0.3265 | 0.9711 | 0.4940 | 0.3328 | **1.0000** | 0.4932 | 0.3327 | 0.4979 |
| 7 | 0.0000 | **0.9578** | 0.0068 | 0.3382 | 0.3486 | 0.3622 | 0.0000 | 0.0000 | 0.5301 | 0.6474 |
| 8 | **0.9608** | 0.8730 | 0.5538 | 0.4027 | 0.3444 | 0.4729 | 0.6258 | 0.2320 | 0.1645 | 0.7588 |
| 9 | 0.9990 | 0.0000 | 0.9989 | 0.0000 | 0.1683 | 0.3702 | 0.0000 | 0.4538 | 0.0847 | 0.0000 |
| OA (%) | 0.5237 | 0.4359 | 0.4376 | 0.5519 | 0.4214 | 0.4207 | 0.5328 | 0.5436 | 0.6010 | **0.6365** |
| Kappa | 0.5246 | 0.4327 | 0.4261 | 0.4570 | 0.3189 | 0.3185 | 0.4536 | 0.4139 | 0.4866 | **0.5429** |
| NMI | **0.5529** | 0.5126 | 0.4758 | 0.4392 | 0.4355 | 0.4362 | 0.5372 | 0.4628 | 0.5142 | 0.5251 |
| ARI | 0.3188 | 0.2632 | 0.2783 | 0.3869 | 0.2691 | 0.2399 | 0.3780 | 0.3271 | 0.5049 | **0.5128** |
| Purity | **0.7012** | 0.6785 | 0.6487 | 0.6638 | 0.5966 | 0.6400 | 0.7126 | 0.6689 | 0.6842 | 0.6980 |

TABLE VI
QUANTITATIVE EXPERIMENTAL CLASSIFICATION RESULTS ON TRENTO,

| No. | k-means | FCM | PCM | NCSC | DFCN | SDCN | EGAE | AdaGAE | DAEGC | AHSGC |
|---|---|---|---|---|---|---|---|---|---|---|
| 1 | 0.0000 | 0.0000 | 0.3673 | **1.0000** | 0.6844 | 0.2218 | 0.6522 | **1.0000** | **1.0000** | 0.9995 |
| 2 | 0.0189 | 0.0000 | 0.0000 | 0.0793 | 0.4191 | 0.7826 | 0.6029 | 0.4826 | 0.1151 | **1.0000** |
| 3 | 0.0000 | 0.0000 | 0.0000 | 0.0000 | 0.0063 | 0.0359 | **0.0931** | 0.0000 | 0.0000 | 0.0000 |
| 4 | 0.7200 | 0.6826 | 0.4502 | **1.0000** | 0.9745 | 0.6237 | 0.9540 | 0.9622 | **1.0000** | **1.0000** |
| 5 | 0.6182 | 0.2007 | 0.4292 | 0.6437 | 0.7105 | 0.7452 | 0.5196 | 0.6158 | 0.6156 | **0.9394** |
| 6 | 0.0199 | 0.0329 | 0.0006 | 0.8206 | 0.3959 | 0.5930 | 0.3412 | 0.8029 | **0.8233** | 0.0224 |
| OA (%) | 0.6301 | 0.5132 | 0.3678 | 0.7489 | 0.7145 | 0.6200 | 0.6487 | 0.7562 | 0.7469 | **0.8603** |
| Kappa | 0.6026 | 0.4917 | 0.3232 | 0.6790 | 0.6314 | 0.5132 | 0.5380 | 0.6810 | 0.6741 | **0.8148** |
| NMI | 0.4928 | 0.4654 | 0.2898 | 0.7312 | 0.6787 | 0.4860 | 0.6801 | 0.7250 | 0.6767 | **0.8844** |
| ARI | 0.3457 | 0.2802 | 0.1303 | 0.7111 | 0.6795 | 0.4362 | 0.6624 | 0.6455 | 0.6483 | **0.8884** |
| Purity | 0.6499 | 0.6708 | 0.5146 | 0.8652 | 0.7817 | 0.7050 | 0.8273 | 0.7726 | 0.8382 | **0.8810** |

*2) Quantitative analysis of the PU dataset:* Compared to the SA dataset, the land cover types in the PU dataset are more dispersed, posing significant challenges for clustering. From the clustering results in Table V, it can also be concluded that the clustering results of all comparison methods have decreased. However, our AHSGC method still performs the best among all investigated clustering methods. Notably, compared with the best results in the comparative methods, the performances of AHSGC are improved by 3.55% in OA, 1.83% in $\kappa$, and 0.77% in ARI, respectively. In addition, we note that deep and graph-based clustering methods do not have significant advantages over traditional clustering methods, mainly because the land features in the PU dataset are scattered and closely related, making it difficult for clustering methods to distinguish. Therefore, it is more important to extract more significant and representative features. It is worth noting that DAEGC and AHSGC have achieved breakthrough performance with an OA of 60.10% and 63.65%, respectively. This is mainly because these two methods adopt the structural graph learning mechanism, which can effectively learn the structural information of the graph, that is, have a better understanding of the latent spatial structural information of the graph.

*3) Quantitative analysis of the Trento dataset:* Compared with the SA and PU datasets, the Trento has a relatively simple distribution of land cover types. However, in Table VI, the performance of traditional clustering methods, i.e., k-means, FCM, and PCM, is still unsatisfactory, the OAs are only 63.01%, 51.32%, and 36.78%, respectively, owing to their limited feature extraction capability. Notably, AHSGC has achieved commendable results, with an OA of 86.03%, exceeding the second-best performance of 10.41%, which is mainly due to its robust powerful feature extraction and discriminative ability between different types of clusters. Due to the homophily-enhanced structure learning module, AHSGC can estimate the interrelationships between nodes and classes, which enables automatic updating of edge connections between nodes, thus improving clustering accuracy. Additionally, the self-supervised structure and feature loss achieve self-representation feature extraction of HSI.

*D. Visual Clustering Results Analysis*

To visually display the clustering results, different colormaps are adopted to represent the clustering results of different land features, as shown in Fig. 5-7. From the visualized clustering results, we can conclude that the clustering results of traditional methods, including K-means, FCM, and PCM, contain a lot of salt and pepper noise, indicating insufficient feature extraction capability and poor noise robustness. Benefiting from their nonlinear representation and high-level feature extraction capabilities, deep learning methods achieve smoother colormaps with less noise. However, for specific details such as edge pixels, misclassification may occur. Due to the adaptive filter, AHSGC can effectively capture the low-pass and high-pass information for clustering, and the excess high-frequency and low-frequency noises are filtered out. In addition, the homophily-enhanced structure learning module, by estimating the edge connections between nodes and classes, can update the connections between nodes and connections between classes, with which the perceptual representation of the HSI topology is effectively captured.

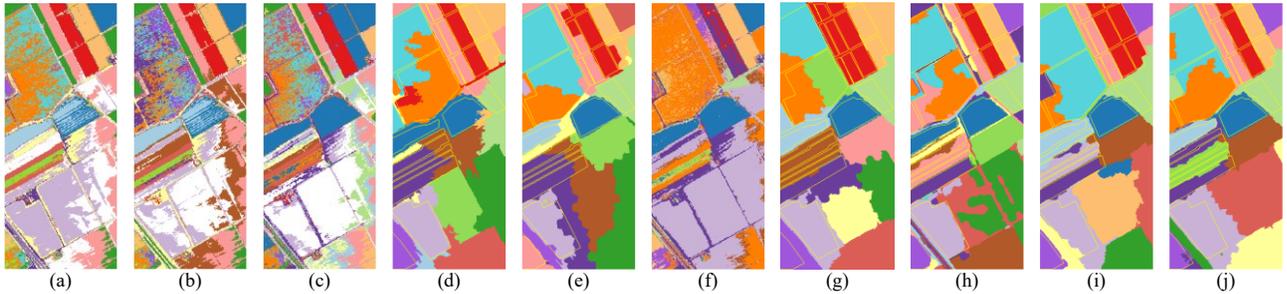

Fig. 5. Clustering maps of SA dataset. (a) K-means (b) FCM (c) PCM (d) NCSC (e) DFCN (f) SDCN (g) EGAE (h) AdaGAE (i) DAEGC (j) AHSGC.

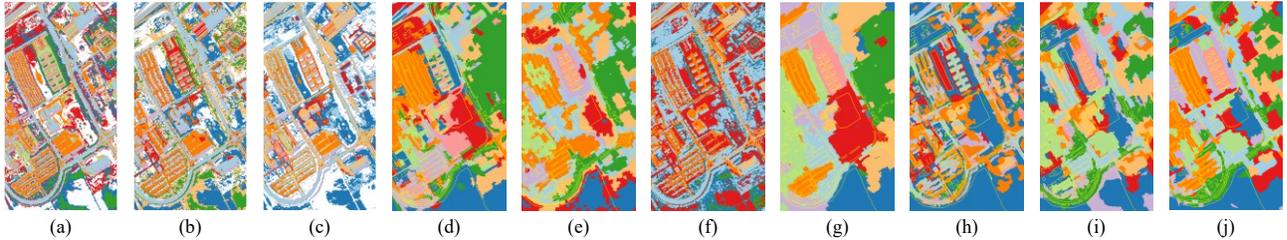

Fig. 6. Clustering maps of PU dataset. (a) K-means (b) FCM (c) PCM (d) NCSC (e) DFCN (f) SDCN (g) EGAE (h) AdaGAE (i) DAEGC (j) AHSGC.

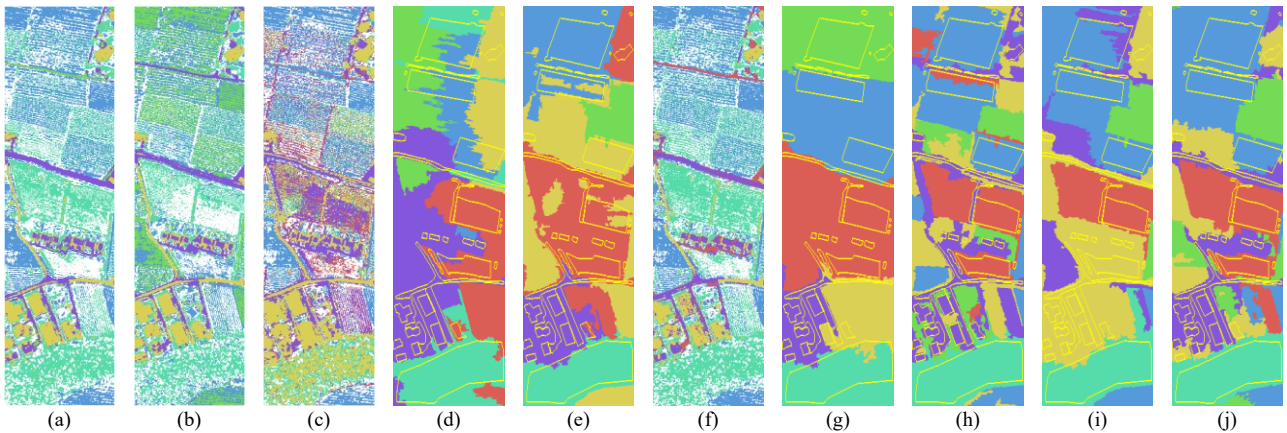

Fig. 7. Clustering maps of Trento dataset. (a) K-means (b) FCM (c) PCM (d) NCSC (e) DFCN (f) SDCN (g) EGAE (h) AdaGAE (i) DAEGC (j) AHSGC.

*E. Complexity Comparison*

In this part, we estimate the runtime and model complexity of all investigated deep clustering methods. We record the indexes of training time, testing time, and FLOP to compare the computational cost and model complexity of different methods. According to the results in Table VII, we note that the training, testing time, and model complexity of our method are much lower than other investigated methods, verifying the success of HSI processing. Additionally, due to AHSGC's powerful and fast feature extraction capabilities, requiring only 50 epochs to complete convergence, its training time is relatively short. Furthermore, AHSGC adopts task-oriented self-training and graph structure reconstruction objective loss to accelerate the ability effectively, and the representative features and the computational efficiency of the model are extracted and enhanced, respectively. Therefore, we can conclude that our method achieves remarkable clustering results with a slight model complexity, which indicates that AHSGC contains excellent practical application prospects.

**TABLE VII**
THE RUNNING TIME AND MODEL COMPLEXITY OF ALL DEEP CLUSTERING METHOS ON THREE DATASETS.

| Name | Salinas | | | PU | | | Trento | | |
|---|---|---|---|---|---|---|---|---|---|
| | Train/s | Test/s | FLOPs | Train/s | Test/s | FLOPs | Train/s | Test/s | FLOPs |
| AdaGAE | 144.8671 | 43.5324 | 130.02M | 1200.819 | 101.3368 | 1.29G | 86.2017 | 13.7605 | 25.53M |
| EAGE | 199.946 | 4.8399 | 270.38M | 1763.686 | 39.8889 | 2.63G | 186.1897 | 4.7581 | 57.09M |
| NCSC | 681.0274 | 15.6768 | 303.18G | 1375.568 | 33.2784 | 596.49G | 677.5501 | 19.3583 | 273.67G |
| DAEGC | 73.4302 | 5.1500 | 148.72 M | 581.7761 | 44.8480 | 1.46 G | 65.9399 | 4.7916 | 27.75 M |
| DFCN | 672.1873 | 4.2803 | 490.69 M | 1802.5410 | 8.5483 | 1.70 G | 282.6037 | 4.4386 | 83.27 M |
| SDCN | 437.5881 | 0.6087 | 446.1G | 274.8834 | 0.2034 | 339.56G | 145.6749 | 0.3492 | 236.22G |
| AHSGC | 179.9025 | 0.0025 | 3.57 G | 237.5779 | 0.0007 | 30.62G | 56.2532 | 0.0629 | 2.95 G |

*F. Parameter Analysis*

In this section, the impacts of different hyperparameters contained in AHSGC on clustering performance are investigated. The number of iterations $T$, learning rate $L$, intra-cluster edge recovery ratio $\xi$, and inter-cluster edge removal ratio $\eta$ are explored through a grid search strategy. Fig.8 and Fig. 9 demonstrate the variations in OA on three datasets with different values of $(\xi, \eta)$ and $(T, L)$, respectively. From the results, we can observe that four paraments achieve a significant impact on finally clustering accuracy, and the model is more likely to achieve superior performance when the $\xi \in [0.5, 0.7]$, $\eta \in [0.3, 0.5]$, $T \in [40, 60]$, and $L \in [3e^{-4}, 5e^{-4}]$, respectively.

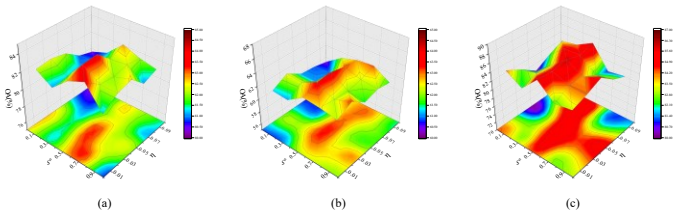

Fig. 8. Sensitivities of $\xi$ and $\eta$ on SA(a), PU(b), and Trento(c).

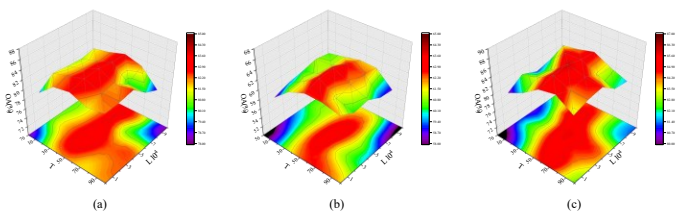

Fig. 9. Sensitivities of $T$ and $L$ on SA(a), PU(b), and Trento(c).

*G. Clustering Visualization*

In this section, the t-distributed stochastic neighbor embedding (*t*-SNE) [42] method is employed to visualize the distribution of graph nodes in three datasets to validate the clustering performance of AHSGC. From Fig.8, (a)-(c) are the original distribution of original graph nodes, while (d)-(f) represent the distribution of graph node features after processing with AHSGC. By comparing and analyzing, the following conclusion can be drawn: 1) the distribution covariance in different classes in (d)-(f) is smaller than that in (a)-(c), indicating that good clustering results have been produced; 2) (d)-(f) demonstrate more considerable inter-class distances and smaller intra-class distances, compared with (a)-(c), showing a more regular and compact distribution; 3) Different land covers are distinguished more clearly. Thus, graph nodes processed by AHSGC exhibit tighter intraclass compactness and more considerable inter-class distances. Furthermore, the clustering task mentioned in Eq. (6) is effectively implemented.

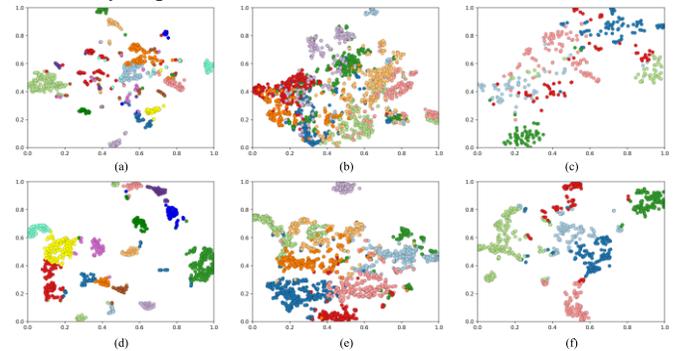

Fig. 8. Visualization of AHSGC using t-SNE on SA (a) and (d), PU (b) and (e), and Trento (c) and (f). (a), (b), and (c) are original distributions of three datasets. (d), (e), and (f) are distributions after clustering. Different color nodes within the maps denote different land covers.

*H. Ablation Experiment*

In this section, we will evaluate the contribution of the adaptive filter graph encoder, homophily-enhanced structure learning, and HSI preprocessing and initial graph construction module to the overall clustering performance. In addition, a set of ablation studies have been conducted. Specifically, AHSGC without the generation of homogeneous regions block is called AHSGC-V1. AHSGC-V2 is formed by removing the adaptive filter graph encoder block. AHSGC-V3 is obtained by removing the homophily-enhanced structure learning block. The results are shown in Fig.9 recording five metrics. From the results, we note that the impacts of different modules in AHSGC on clustering accuracy are different. Additionally, each module in the AHSGC contributes to the improvement of clustering accuracy.

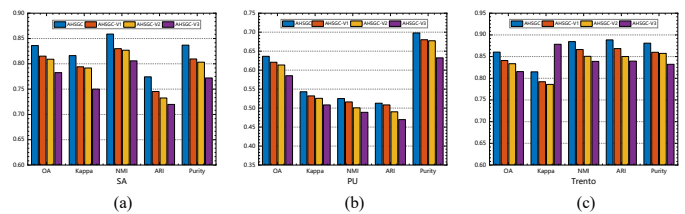

Fig. 8. Ablation experiment results with five metrics on three datasets.

## V. CONCLUSION

This paper proposes a novel homophily enhanced structure graph learning with an adaptive filter clustering method (AHSGC) for HSI. Specifically, we adopt a homogeneous region generation method to preprocess the HSI, with which the local spatial-spectral structure information is preserved, and the node number in the graph is reduced. After that, to adaptively capture the high and low frequency features on the graph for clustering, an adaptive filter graph encoder is introduced. Then, a graph embedding clustering self-training decoder is developed with KL Divergence, with which the pseudo-label is generated for network training. Meanwhile, we design a homophily enhanced structure graph learning module to update the graph according to the task, in which the orient correlation estimation is adopted to estimate the node connection, and graph edge sparsification is designed to adjust the edges in the graph dynamically. Finally, the joint network optimization is proposed to optimize the network, and the K-means is adopted to express the latent features.

In future work, some more spatial-spectral self-supervised methods will be explored for graph learning. In addition, some more research, e.g., reinforcement learning, will be applied for HSI preprocessing to enhance the feature extraction abilities of clustering methods.

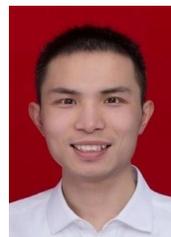

**Yao Ding** received the M.S. and Ph.D. degree from the Key Laboratory of Optical Engineering, Xi'an Research Institute of High Technology, Xi'an 710025, China, in 2013 and 2022. His research interests include neural network, computer vision, image processing, and hyperspectral image clustering. He has published several papers in IEEE Trans. on Geoscience and Remote Sensing (TGRS), Information Sciences (INS), Expert Systems with Applications (ESWA), Defence Technology (DT), IEEE Geoscience and Remote Sensing Letters (GRSL), Neurocomputing, etc. Furthermore, he has published three monographs, and six patents have been applied. He has received excellent doctoral dissertations from the China Simulation Society and the China Ordnance Industry Society in 2023. He also has received HIGHLY CITED AWARDS from Defence Technology (DT) journal. At present, he has Eleven highly cited papers of ESI. In addition, he is also the reviewer of TGRS, TNNLS, PR, JAG, KBS, etc. He has also served as a Youth editorial board member of Journal of Information and Intelligence.

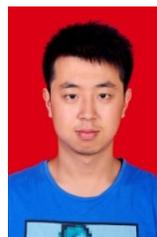

**Weijie Kang** is a lecturer at the Xi'an-Research Institute of HighTechnology, China. He received his B.S. degree in 2017 from Xi'an Jiaotong University, received his M.S., and Ph.D. degrees in 2019, 2022 from Air Force Engineering University. His main research interest is adaptive signal processing, system modeling, and health management.

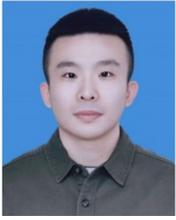
**Aitao Yang** received the B.S. degree in information engineering from Xi'an Institute of High Technology, Xi'an, China, in 2017. He is currently pursuing the Ph.D. degree in computer science and technology at Xi'an Hi-Tech Research Institute, in 2021.

His main research interests include remote sensing information processing, hyperspectral image clustering, and computer vision.

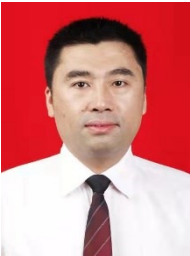
**Zhili Zhang** received the B.S., M.S., and Ph.D. degrees in 1988, 1991, and 2001, respectively, all from Xi'an Research Institute of Hi-Tech, China. He is currently a professor in Xi'an Research of Hi-Tech, China. His research interests include inertial navigation systems theory, system simulation, image processing and Position and navigation.

He has achieved fruitful results in the field he is studying, and several Prizes have been awarded, including:1) One First Class Prize and Two Second Class Prizes of The State Science and Technology Progress Award of China; 2) One Second Class Prize of The State Technology Invention Award of China; 3) Twelve Provincial/Ministerial level awards. In addition, he has published six monographs, four briefs, and over a hundred SCI and EI research papers, including four hot papers and ten highly cited papers of ESI. Furthermore, he has been authorized over fifty invention patents.